\documentclass[12pt]{article}
\usepackage{amsmath,amssymb}
\usepackage{graphicx}
\usepackage{enumerate}
\usepackage{natbib}
\usepackage{url} 

\newtheorem{theorem}{Theorem}
\newcommand{\E}{\mathbb{E}}

\renewcommand{\P}{\mathbb{P}}

\begin{document}

\bibliographystyle{natbib}

\def\spacingset#1{\renewcommand{\baselinestretch}%
{#1}\small\normalsize} \spacingset{1}


\begin{center}
{\LARGE\bf Forest Guided Smoothing}\\
Isabella Verdinelli and Larry Wasserman\\
Department of Statistics and Data Science\\
Carnegie Mellon University
\end{center}

\begin{center}
March 8 2021
\end{center}

\bigskip
\begin{abstract}
We use the output
of a random forest
to define a family of
local smoothers
with spatially adaptive
bandwidth matrices.
The smoother inherits
the flexibility of the
original forest but,
since it is a simple, linear smoother,
it is very interpretable and
it can be used for tasks that
would be intractable for the
original forest.
This includes
bias correction,
confidence intervals,
assessing variable importance
and methods for exploring the structure of the forest.
We illustrate the method
on some synthetic examples and
on data related to Covid-19.

\end{abstract}

\noindent%
\vfill

\newpage
\spacingset{1.5} 
\section{Introduction}
\label{sec:intro}

Random forests are often an
accurate method for nonparametric regression
but they are notoriously
difficult to interpret.
Also, it is difficult to construct 
standard errors,
confidence intervals
and meaningful measures of variable importance.
In this paper,
we construct
a spatially adaptive local linear smoother
that approximates the forest.
Our approach builds on the ideas in
\cite{bloniarz2016supervised}
and
\cite{friedberg2020local}.
The main difference is that we define a 
one parameter family
of bandwidth matrices
which help
with the construction of confidence intervals,
and measures of variable importance.

Our starting point is the well-known
fact that
a random forest can be regarded
as a type of 
kernel smoother
(\cite{breiman2000randomizing,
scornet2016random,
lin2006random,
geurts2006extremely,
hothorn2004bagging,meinshausen2006quantile}).
We take it as a given that
the forest is an accurate predictor
and we do not make any attempt to
improve the method.
Instead,
we want to find a family of linear smoothers
that approximate the forest.
Then we show how to use
this family
for
interpretation,
bias correction,
confidence intervals,
variable importance
and for exploring the structure of the forest.

\bigskip

{\bf Related Work.}
Our work builds on
\cite{bloniarz2016supervised} and
\cite{friedberg2020local}.
\cite{bloniarz2016supervised} 
fit a local linear regression using weights
from a random forest.
They show that this often leads to improved prediction.
\cite{friedberg2020local}
go further and modify the forest algorithm
to account for the fact that a local linear fit will be
used and to reduce the bias of the fit.
This further improves the performance and yields confidence intervals.

We use the forest weights
to fit a local linear regression
but we do so by first
building a 
family of bandwidth matrices
$\{ h H_x:\ h>0, x\in \mathbb{R}^d\}$
depending on one free parameter $h>0$.
We use the bandwidth matrices
to define a kernel from which we 
get the local linear fit.
Creating the bandwidth matrices has several advantages.
First, it allows us to use
the generalized jackknife to correct the bias
and construct confidence intervals.
In contrast to 
\cite{friedberg2020local},
this allows us to use any off-the-shelf random forest;
no adjustments to the forest algorithm are required.
Second, the collection of bandwidth matrices
will be used to create several summaries
of the forest. For example, we can examine how much smoothing 
is done with respect to different covariates and in different 
parts of the covariate space.
We also define the notion of a typical bandwidth matrix
using the Wasserstein barycenter.
Third, we can explore variable importance
based on local slopes at different resolutions by varying the 
parameter $h$ 
thus giving a multiresolution measure of variable importance.

\bigskip

{\bf Paper Outline.}
In Section \ref{section::fgs}
we define the forest guided smoother.
In Section \ref{section::confidence}
we discuss the construction of
confidence intervals.
In Section \ref{section::explore}
we present methods for
exploring the structure of the forest.
Examples are presented in Section \ref{section::examples}.
Section \ref{section::discussion}
contains concluding remarks.

\section{Forest-Guided Smoothers}
\label{section::fgs}

Let
$$
(X_1,Y_1),\ldots, (X_n,Y_n)\sim P
$$
where
$Y_i\in\mathbb{R}$ and
$X_i\in\mathbb{R}^d$.
We assume that $d < n$ and is fixed.
Let
$\mu(x) = \mathbb{E}[Y|X=x]$
denote the regression function.
Recall that the random forest estimator
$\hat \mu_{RF}(x)$ is
$$
\hat\mu_{RF}(x) = \frac{1}{B}\sum_{j=1}^B \hat\mu_j(x)
$$
where
each $\hat\mu_j$ is a tree estimator
built from a random subsample of the data,
a random subsample of features
and $B$ is the number of subsamples.

We take, as a starting point,
the assumption that 
$\hat\mu_{RF}$ is a good estimator.
Our goal is not to improve
the random forest
or provide explanations
for its success.
Rather, we construct an estimator
that provides a tractable approximation
to the forest which can then be used
for other tasks.

As noted by
\cite{hothorn2004bagging,
meinshausen2006quantile}
the random forest estimator $\hat\mu(x)_{RF}$ can be re-written as
$$
\hat \mu_{RF}(x) =\sum_{i=1}^n w_i(x) Y_i
$$
for some weights
$w_i(x)$
where $w_i(x) \geq 0$ and
$\sum_i w_i(x)=1$.
As these authors note,
these weights behave like a 
spatially adaptive kernel.

We proceed as follows.
As in 
\cite{friedberg2020local}
we split the data into two groups
$D_1$ and $D_2$.
For simplicity, assume each has size $n$.
We construct a random forest $\hat\mu_{RF}$ from $D_1$.
Now we define the bandwidth matrix
\begin{equation}\label{eq::Hx}
H_x =\left( \frac{1}{n}\sum_i w_i(x) (X_i-x)(X_i-x)^T\right)^{1/2}
\end{equation}
where the sum is over $D_1$.
Let $K$ be a spherically symmetric kernel and define
$$
K(x;H_x) = |H_x|^{-1}K(H_x^{-1}x).
$$
This yields a kernel centered at $x$
whose scale matches the scale of the forest weights.
We then define
the one parameter family of bandwidth matrices
$\Xi=\{h H_x:\ h>0, x\in\mathbb{R}^d\}$.

We define the forest guided local linear smoother, or FGS,
to be the local linear smoother $\hat\mu_h(x)$ with
kernel $K(x;h H_x)$, that is,
$\hat\mu_h(x) = \hat\beta_0(x)$ obtained by minimizing
$$
\sum_i \Bigl(Y_i -\beta_0(x) - \beta(x)^T
            (X_i-x)\Bigr)^2 K(X_i-x;hH_x).
$$
Then
$$
\hat\mu_h(x) = 
e_1^T (X_x^T W_x X_x)^{-1} X_x W_x Y = \sum_i \ell_i(x;hH_x) Y_i
$$
where
$$
X_x=
\left[
\begin{array}{cc}
1 & (X_1-x)^T\\
\vdots & \vdots\\
1 & (X_n - x)^T
\end{array}
\right],
$$
$W_{x}$ is a diagonal matrix with $W_{x}(i,i) = K(X_i- x;hH_x)$,
$e_1 = (1,0,\ldots, 0)^T$ and
\begin{equation}\label{eq::ell}
\ell(x;hH_x)= e_1^T (X_x^T W_x X_x)^{-1} X_x W_x.
\end{equation}
When $h=1$, which can be regarded as a default value,
we write $\hat\mu_h(x)$ simply as $\hat\mu(x)$.
Although we focus
on local linear regression,
one can also use this 
for kernel regression or higher order local polynomial regression.
We shall see that $\hat\mu(x)$ is often
a good approximation to $\hat\mu_{RF}(x)$.

\bigskip

{\bf Remark:}
Other 
approaches for choosing $H_x$ are possible.
For example, one could minimize the difference
between $K(x-X_i;H_x)$ and $w_i(x)$ over
all positive definite matrices $H_x$.
However,
(\ref{eq::Hx}) is simple
and in our experience 
works quite well.
In high dimensional cases,
$H_x$ would require regularization
but we do not pursue 
the high dimensional case in this paper.

\bigskip

\begin{figure}
\begin{center}
\includegraphics[scale=.5]{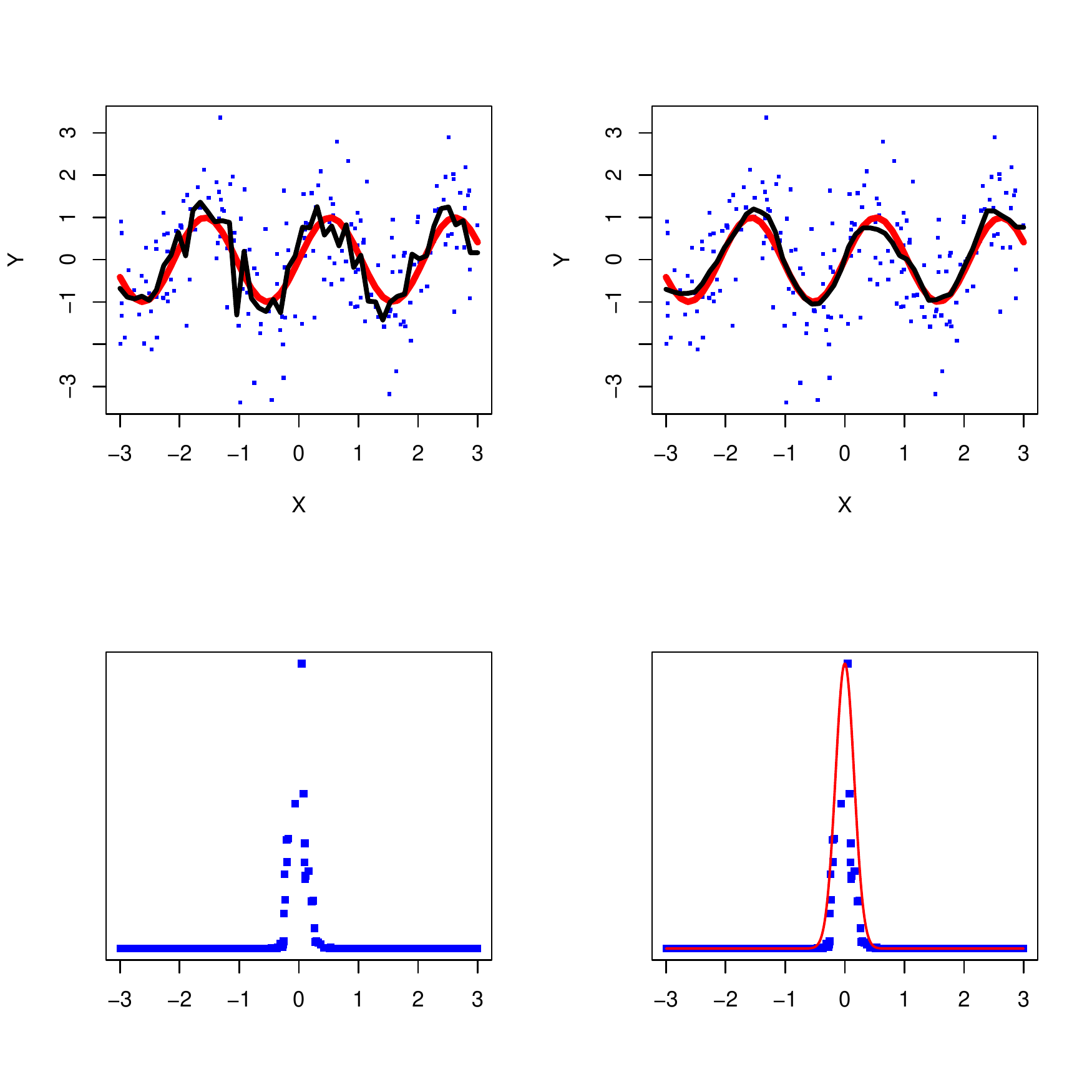}
\end{center}
\vspace{-1.5cm}
\caption{\small
Top left:
The data points, the random forest estimator $\hat\mu_{RF}(x)$ in 
black, and the true function in red.
Top right: forest guided smoother $\hat\mu(x)$ in black,
and the true function in red.
Bottom left: forest weights
$w_1(x),\ldots, w_n(x)$ evaluated at $x=0$.
Bottom right:
kernel
approximation to the weights.}
\label{fig::Cool1a}
\end{figure}

\begin{figure}
\begin{center}
\includegraphics[scale=.4]{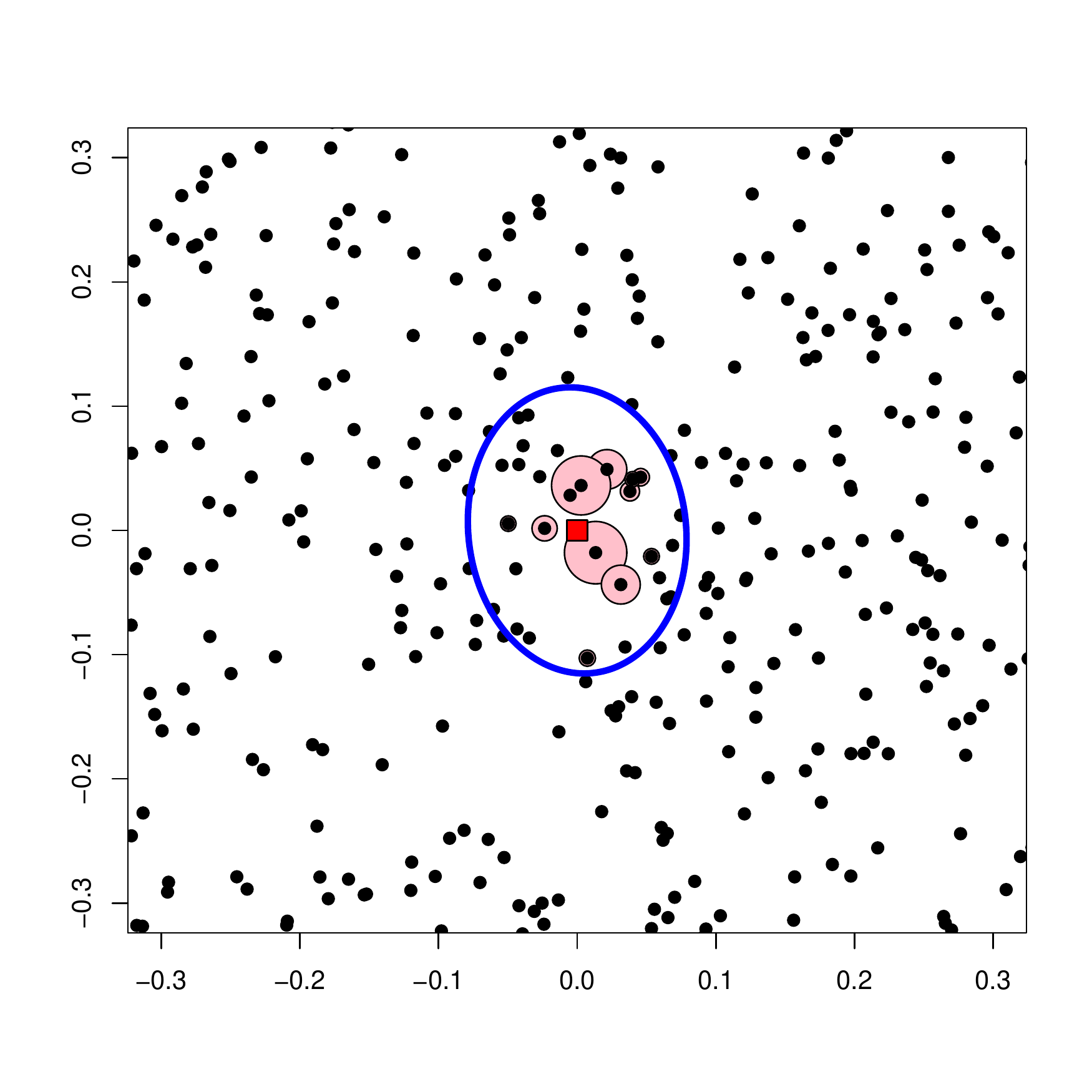}
\end{center}
\vspace{-1.5cm}
\caption{\small The dots represent data point.
The target point $x$ is indicated by the red square.
The gray circles show the forest weights
and the blue ellipse represents the kernel
approximation to the weights.}
\label{fig::Cool1b}
\end{figure}

Figure \ref{fig::Cool1a}
shows a one-dimensional example.
The top left shows
the data, the random forest estimator $\hat\mu_{RF}$ in lack,
and the true function in red.
The forest guided smoother $\hat\mu(x)$ is the black line in the 
top right plot.
The bottom left shows the weights
$w_1(x),\ldots, w_n(x)$ 
at $x=0$
and the
the bottom right shows our kernel
approximation to the weights.
We see that the 
FGS approximates the forest and
the kernel approximates the weights very well.
Figure \ref{fig::Cool1b}
shows a two-dimensional example.
Here we show the forest weights as gray circles
and the ellipse represents the
approximating kernel.
The target point is (0,0).
Again, the kernel approximates the
weights.

For getting standard
errors and confidence intervals,
we will also need
to estimate
the variance
$$
\sigma^2(x) = {\rm Var}(Y|X=x).
$$
We will proceed as follows.
Let $r_i = Y_i - \hat\mu_{RF}(X_i)$ be the residuals from the forest.
We regress the $r_i^2$'s on $X_i$'s
to estimate $\sigma^2(x)$
using another random forest.
We find that this approach tends to under-estimate $\sigma^2(x)$
in some cases and we replace
$\hat\sigma(x)$ with 
$c\,\hat\sigma(x)$ where we use $c=1.5$
as a default to compensate for this
in our examples.

\vspace{-.5cm}
\section{Confidence Intervals}
\label{section::confidence}

In this section 
we construct
estimators 
of the bias of $\hat\mu(x)$
and then obtain
confidence intervals for $\mu(x)$.
This is difficult to do
directly from the forest
without delicate modifications of the forest algorithm
to undersmooth,
as in
\cite{friedberg2020local}.
But bias estimation using standard methods
is possible with the FGS.
We start by recalling
some basic properties
of local linear smoothers.

\vspace{-.5cm}
\subsection{Properties of Smoothers}

Let $\hat\mu$
be the local linear smoother 
based on bandwidth matrices
$H_x \equiv H_{n,x}$.
Let
$f(x)$ be the density of $X$,
define $\mu_2(K)$ by
$\int u u^T K(u)du= \mu_2(K)I$,
and $R(K) = \int K^2(u) du$.
Let ${\rm Hess}$ be the Hessian of $\mu$.
\cite{ruppert1994multivariate}
consider the following assumptions.

\smallskip

(A1) $K$ is compactly supported and bounded.
All odd moments of $K$ vanish.

\smallskip

(A2) $\sigma^2(x)$ is continuous at $x$
and $f$ is continuously differentiable.
Also, the second order derivatives of $\mu$
are continuous.
Further,
$f(x)>0$ and $\sigma^2(x)>0$.

\smallskip

(A3) 
$H_{n,x}$ is symmetric and positive definite.
As $n\to\infty$ we have
$n^{-1}|H_{n,x}|\to 0$
and
$H_{n,x}(i,j)\to 0$
for every $i$ and $j$.

\smallskip

(A4)
There exists $c_\lambda$ such that
$$
\frac{\lambda_{\rm max}(H_{n,x})}{\lambda_{\rm min}(H_{n,x})}\leq c_\lambda
$$
for all $n$
where $\lambda_{\rm max}$ and $\lambda_{\rm min}$ 
denote the maximum and minimum eigenvalues.

\bigskip

Under these conditions, 
\cite{ruppert1994multivariate}
showed that the bias 
$B(x,H_x)$
and variance 
$V(x,H_x)$
of
$\hat\mu(x)$,
conditional on $X_1,\ldots, X_n$
are
\begin{equation}\label{eq::bias}
B(x,H_x)=
\frac{1}{2}\mu_2(K)
{\rm tr}(H_x^2 {\rm Hess}(x)) + o_P( {\rm tr}(H_x^2))
\end{equation}
and
\begin{equation}\label{eq::var}
V(x,H_x) =
\frac{\sigma^2(x)R(K)}{n |H_x| f(x)} ( 1 + o_P(1)).
\end{equation}
It follows that the bias using bandwidth $h H_x$ satisfies
$$
B(x,hH_x) = h^2 c_n(x) + o_P(h^2 {\rm tr}(H_x^2))
$$
for some $c_n(x)$.

\bigskip

Assumptions (A3) and (A4) capture the idea
that the bandwidth matrix needs to shrink towards 0 in some sense.
Assumption (A4) essentially says that
$H_{n,x}$ behaves like a scalar tending to 0
times a fixed positive definite matrix.
For our results,
we will make this more explicit and slightly strengthen
(A4) to:

\smallskip

(A4) There exists a sequence $\phi_n\to 0$ and a positive definite symmetric matrix
$C_x$ such that
$H_{n,x} \sim \phi_n C(x)$
where
$\phi_n\asymp (1/n)^{a}$
for some $0 < a < 1$.

\smallskip

With (A4) we have
$B(x,hH_x) = h^2 c(x)/n^{2} + o_P(h^2)$.
To construct the bias correction we 
need to add the following stronger smoothness condition.

\smallskip

(A5)
For some $t$,
the $t^{\rm th}$ order derivatives of $\mu$ are continuous and
there exist
functions
$c_1(x),\ldots, c_t(x)$ such that,
for any $h>0$,
$$
B(x,h H_x) = \sum_{j=2}^t \frac{c_j(x)h^{j}}{n^{aj}} + o_P\left( \frac{1}{n^{at}}\right).
$$

\smallskip

\cite{ruppert1997empirical}
showed
how to estimate the bias
of $\hat\mu(x)$
by fitting the estimator
for several values of the bandwidth.
This type of bias estimation
has been used in other contexts and
is sometimes referred to as
generalized jackknife;
see for example \cite{cattaneo2013generalized}.

\bigskip

In more detail,
Ruppert's method (i.e. the generalized jackknife) works as follows.
Choose a set of $b$ bandwidths 
$h_1,h_2,\ldots, h_b$ and let
$\hat m = (\hat \mu_{h_1}(x),\ldots, \hat\mu_{h_b}(x))$.
Let
$\kappa_n = (\mu(x),\kappa_{2,n}(x),\ldots, \kappa_{t,n}(x))^T$
where
$\kappa_{j,n}(x) = c_j(x) /n^{ja}$.
Let
$$
{\cal H}=
       \begin{bmatrix}
        1 \     & h_1^2 \ &  h_1^3 \!\!& \ldots h_{1}^t \\
\vspace{-.4cm}\\
        1 \     & h_2^2 \ & h_2^3 \!\!& \ldots h_{2}^t \\
        \vdots  & \vdots  & \vdots    &  \vdots\\
        1 \ & h_b^2 \ & h_b^4 \!\!& \ldots h_{b}^t \\
          \end{bmatrix} .
$$
We estimate $\kappa_n$ by least squares, namely,
$$
\hat \kappa_n = 
{\rm argmin}_c \ ||\hat m - {\cal H}\,c||^2 =
({\cal H}^T{\cal H})^{-1}{\cal H}^T \hat m.
$$
Now
$\hat m = L Y$ where
$$
L = 
   \begin{bmatrix} 
   \ell_{1}(x;h_1H_x) & \ell_{2}(x;h_1H_x) & \ldots  
                       \ell_{n}(x;h_1H_x) \\
\vspace{-.4cm}\\
 \ell_{1}(x;h_2H_x) &    \ell_{2}(x;h_2 H_x) & \ldots
                        \ell_{n}(x;h_2 H_x) \\ 
 \vdots  & \quad \vdots & \quad \vdots  \\
 \ell_{1}(x;h_b H_x) & \ell_{2}(x;h_b H_x) & \ldots  
                      \ell_{n}(x;h_b H_x) \\ 
   \end{bmatrix}
$$
where $\ell_i(x;h_j H_x)$
are the elements of the vector
$\ell(x;h_j H_x)$ defined in (\ref{eq::ell}).
Therefore
$$
\hat \kappa_n = ({\cal H}^T{\cal H})^{-1}{\cal H}^T L\ Y.
$$
We estimate the bias of $\hat\mu_h(x)$ by
$$
\hat B(x, h) = \sum_{j=2}^t \hat\kappa_{j,n}(x) h^{j} =
g^T ({\cal H}^T{\cal H})^{-1}{\cal H}^T L\ Y
$$
where $g = (0, h^2, \ldots, h^t)^T$. 
The de-biased estimator is
the first element of $\hat\kappa_n$, that is,
$$
\mu^\dagger(x) = e_1^T ({\cal H}^T{\cal H})^{-1}{\cal H}^T L Y =
\sum_i \tilde{\ell}_i(x)
$$
where
$\tilde{\ell}(x)= e_1^T ({\cal H}^T{\cal H})^{-1}{\cal H}^T L$.

The variance of $\mu^\dagger(x)$ 
(conditional on the $X_i$'s) is
$$
{\rm Var}[\mu^\dagger(x)] =
\sum_i \tilde{\ell}_i^2(x) \sigma^2(X_i)
$$
and the estimated variance is
$$
s^2(x)  = \sum_i \tilde{\ell}_i^2(x) \hat\sigma^2(X_i).
$$

Ruppert used the bias estimation method
as part of a bandwidth selection method.
We are interested, instead, to 
get a centered
central limit theorem.
We now confirm that this indeed works.
For the theory,
we need to be more specific about the choice
of bandwidths in the bias correction procedure.
Specifically,
let 
$h_j= \alpha_j n^{-\gamma}, \;{\rm for}\;j=1,2,\ldots b$, with
$0<\alpha_1<\ldots<~\alpha_b$
being constants not depending on $n$.

\begin{theorem}
\label{thm::clt}
Assume that,
conditional on $D_1$, assumptions (A1)-(A5) hold and:
\begin{enumerate}
\item[(i)] \qquad $\sup_x | \hat\sigma^2(x) - \sigma^2(x)| 
           \stackrel{P}{\to} 0$,\\
\vspace{-1cm}
\item[(ii)] \qquad $-a < \gamma < \frac{1-ad}{d}$.
\end{enumerate}
Further, if
$t < d/2$ we require
$a < 1/(d-2t)$.
Also, assume that $Y$ is bounded and that $b > t+1$.
Then
$$
\frac{\mu^\dagger(x) - \mu(x)}{s(x)}\rightsquigarrow N(0,1).
$$
Hence,
$$
\P(\mu(x)\in C_n(x))\to 1-\alpha
$$
where $C_n(x) = \mu^\dagger(x) \pm z_{\alpha/2}  s(x)$.
\end{theorem}

The proof is in the appendix.

It is important to note that
$\gamma$ can be 0 or even negative.
\cite{ruppert1997empirical} requires
$\gamma>0$.
The difference is that
our bandwidth is of the form
$h H_x$ and $H_x$ is already
tending to 0 
and we only need the product to go to 0.
This significantly
simplifies the choice of grid of bandwidths
because the bandwidths can be constant order
and don't need to change with $n$.
For example, one could use a grid like
$(1/8,1/4,1/2,1,2,4,8)$.
We recommend including $h=1$ in the grid
as this corresponds to the original FGS.

\subsection{Variability Intervals}

A commonly used alternative to
confidence intervals
for nonparametric regression
is to form some sort of interval around
the estimate that informally represents
uncertainty but without the coverage claim
of a confidence interval.
We will refer to these as variability intervals.
The simplest approach is to
use $\hat\mu(x) \pm c_\alpha s(x)$
where $s^2(x)$ is the estimated variance
of $\hat\mu(x)$.
If $\hat\mu(x)$ satisfies a central limit theorem
and $c_\alpha = z_{\alpha/2}$
then this is a confidence interval for
$\E[\hat\mu(x)]$.

In our case, such a variability interval is simply
$C_n(x) = \hat\mu(x) \pm z_{\alpha/2} s(x)$.
The extra parameter $h$ is not needed
since we do not use the
generalized jackknife to reduce the bias.
However, it might be useful
to construct multiresolution variability intervals
at various resolutions $h$.
This is the approach to inference
recommended by
\cite{chaudhuri2000scale}
who refer to this as scale-space inference.

In Section \ref{section::examples}
we illustrate this multiresolution approach
for estimating the gradient
as a measure of variable importance.

\subsection{Discussion of Other Methods}

Variability intervals
for forests
have been obtained in
\cite{mentch2016quantifying, peng2019asymptotic}
by deriving a U-statistic based
central limit theorem.
\cite{wager2014confidence}
estimate the variance of the forest
using the jackknife.
The advantage of these approaches
is that they do not need to use
sample splitting as we do.

Confidence intervals
were obtained by
\cite{athey2019generalized}
and
\cite{friedberg2020local}.
They also use data splitting.
The main difference is that
we leave the forest algorithm untouched
and we use the generalized jackknife to reduce the bias.
Instead, they modify the construction of the
forest and
require that the forest is constructed
to satisfy certain assumptions; specifically
they require
that the forest is built from
subsamples of size
$s\asymp n^{\beta}$ where
$$
\beta_{\rm min} = 1 -
\left(
1 + \frac{d}{\pi} \frac{\log (1/\alpha)}
                       {\log 1/(1-\alpha)}\right) < \beta < 1,
$$
where
$\pi/d$ is a lower bound
on the probability of splitting on a feature
and each tree leaves
a fraction of points $\alpha$ on each size of every split.

The advantage of this approach is that
it only requires the regression function to be Lipschitz
whereas the generalized jackknife 
assumes that $\mu(x)$ has at least $t+1$ derivatives.
The disadvantage is that the
conditions on the construction of the forest
are rather complicated and non-standard
and one cannot use any off-the-shelf forest.
As noted in
\cite{friedberg2020local},
the tuning of forest parameters in practice
can be quite different than what is assumed in the theory.
Our main assumption 
is simply that the local smoother
has standard bias and variance properties.

Both approaches require assumptions and
it is difficult to say that
one set of assumptions is better than the other
as they are quite incomparable.
One is an assumption about the algorithm and the other
is an assumption about the function and the bandwidth.

\bigskip

{\bf Remark:}
It may be the case that
there are irrelevant variables.
That is, we have have that
$\mu(x) = \mu(x_S)$
for some subset of variables $x_S$.
If the forest is able to
discover the relevant variables,
then the bandwidth matrix $H_x$
might not shrink
in the direction of the irrelevant variables.
This is a good thing but, technically,
the conditions
(A3-A5) may be violated.
However, the gradient and Hessian of $\mu(x)$
vanish in the irrelevant directions
and Theorem \ref{thm::clt} still holds.

\subsection{Examples of Confidence Intervals}

Now we consider some examples.
In each case, we use $t=2$.
The results using $t=3$ and $t=4$
are similar.

\begin{figure}
\begin{center}
\includegraphics[scale=.75]{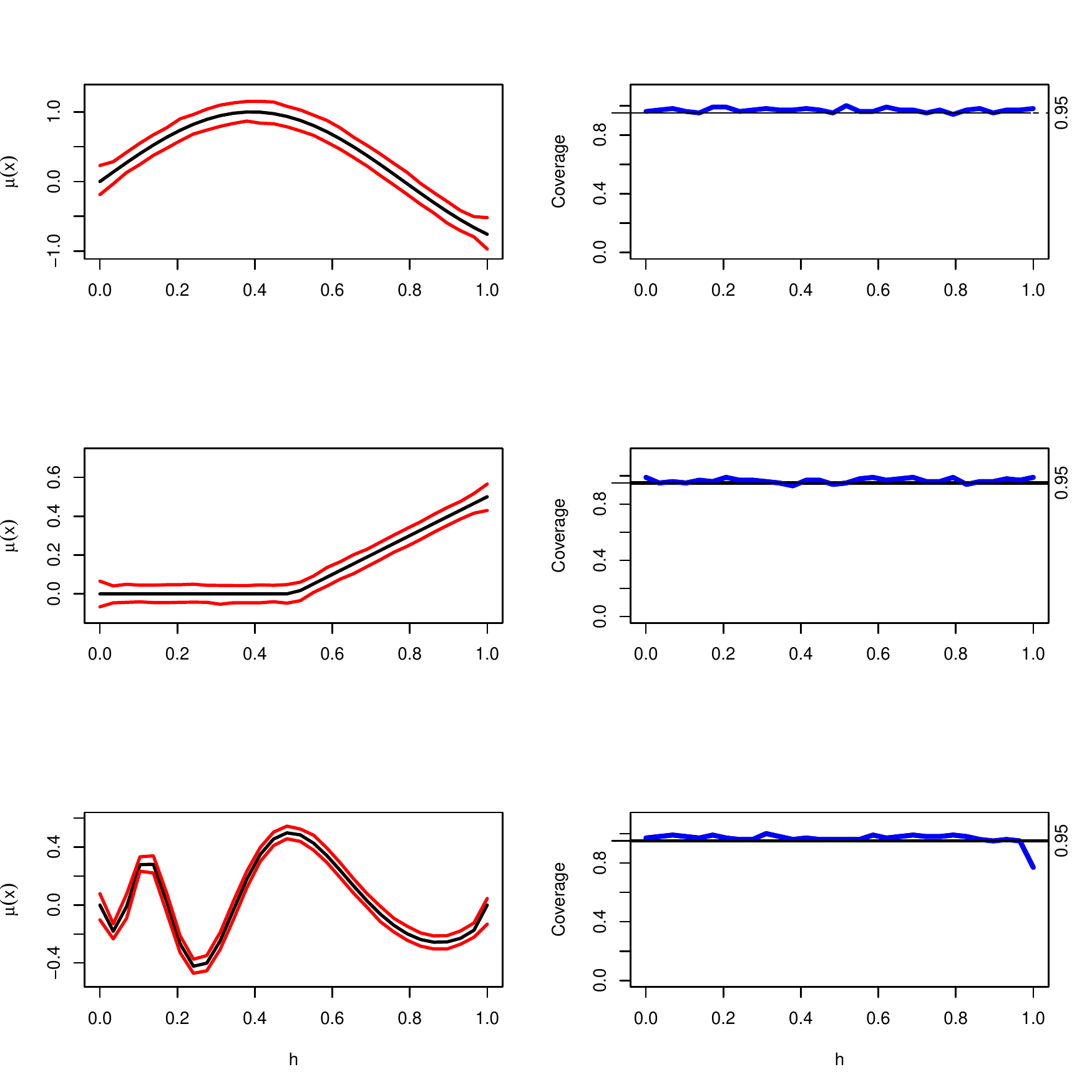}
\end{center}
\caption{\small The figure shows three one-dimensional examples.
The black lines on the left plots show the true function.
The red lines on the left plots show 
the confidence bands from one simulation.
The right plots show the estimated coverage at each $x$ 
based on 100 simulations.}
\label{fig::plot_coverage_one_dimension}
\end{figure}

Figure \ref{fig::plot_coverage_one_dimension}
shows three, one dimensional examples.
The plots on the left
show the true functions in black
and the average over 100 simulations of the pointwise 95 percent
confidence bands in red.
The plots on the right show
the coverage (estimated by simulation)
as a function of $x$.
The sample size in each case is $n=1,000$.
The functions are
$\mu(x) = \sin(4x)$,
$\mu(x) = I(x > 1/2) - 1/2$
and
$\mu(x) = \sqrt{ x(1-x)\sin(2.1 \pi/(x+.35))}$.
The data were generated as
$Y_i = \mu(X_i) + \sigma \epsilon_i$
where
$\epsilon_i\sim N(0,1)$,
$\sigma = .1, .03, .03$ in the three examples
and $X_i \sim {\rm Unif}(0,1)$.
We take the grid of bandwidths
$h_1,\ldots,h_b$ to be an equally spaced grid of size 20
from $h=.1$ to $h=2$.
In each case the coverage reaches its nominal value.

Next we consider 
some multivariate examples.
The first is from 
\cite{friedman1995introduction}
and is
$Y_i = \mu(X_i) + \sigma \epsilon$
where
\begin{equation}\label{eq::mu1}
\mu(x) = 10 \sin (\pi x_1 x_2) + 20(x_3 - 0.5)^2 + 10 x_4 + 5 x_5 
\end{equation}
with $n=500$, $\sigma=1$
and $X_i$ is uniform on $[0,1]^5$.
We take $h$ to be in an equally spaced grid of size 20
from $h=1$ to $h=5$.
We construct 90 percent confidence intervals
at 10 randomly selected points.
The second example is from 
\cite{friedberg2020local}
and is
$Y_i = \mu(X_i) + \sigma \epsilon$
where now
\begin{equation}
\mu(x) = 
\frac{10}{1 + \exp(-10(x_1-.5))} + 
\frac{5}{1 + \exp(-10(x_2-.5))}
\label{eq::mu2}
\end{equation}
with $n=500$, $\sigma=5$
and $X_i$ is uniform on $[0,1]^5$.
We take $h$ to be in an equally spaced grid of size 20
from $h=1$ to $h=30$.
We construct 90 percent confidence intervals
at 10 randomly selected points.

Table ~\ref{table::coverage1} and ~\ref{table::coverage2}
show coverage and average length of confidence intervals
at 10 randomly chosen points for the functions in 
(\ref{eq::mu1}) and (\ref{eq::mu2}). The coverage is close 
to the nominal value
and the lengths are close to those
in \cite{friedman1995introduction} and 
\cite{friedberg2020local}.

\bigskip

{\bf Remark.}
Our grids
were chosen to achieve good coverage
and length for the examples.
In practice we suggest a grid 
ranging from $h=.1$ to $h=10$.
While this choice cannot be claimed to be optimal,
and may not eliminate the bias,
it should result in some amount of bias reduction.
As pointed out in the discussion of
\cite{cattaneo2013generalized},
finding an optimal grid for the generalized jackknife
is an unsolved problem.

\begin{table}
\caption{\small Coverage and average length of confidence 
intervals for the function in \eqref{eq::mu1}}
\begin{center}
\begin{tabular}{l|cccccccccc} \hline\hline
         & 1     & 2    &  3   &   4  & 5    & 6    & 
           7    & 8    & 9    & 10\\  \hline
Coverage & 0.87  & 0.84 & 0.84 & 0.92 & 0.88 & 0.88 & 
           0.85 & 0.90 & 0.85 & 0.86\\
Length   & 4.66  & 4.47 & 4.53 & 4.72 & 4.53 & 4.84 & 
           4.72 & 4.86 & 4.49 & 4.59\\ \hline\hline
\end{tabular}
\end{center}
\label{table::coverage1}
\end{table}

\begin{table}
\caption{\small Coverage and average length of confidence 
intervals for the function in \eqref{eq::mu2}}
\begin{center}
\begin{tabular}{l|cccccccccc} \hline\hline
         & 1     & 2    &  3   &   4   & 5    & 6    & 
           7     & 8    & 9     & 10\\  \hline
Coverage & 0.93  & 0.84 & 0.91 & 0.91  & 0.92 & 0.93 & 
           0.85  & 0.92 & 0.87  & 0.94\\
Length   & 9.63  & 9.62 & 9.61 & 10.14 & 9.39 & 9.98 & 
           10.75 & 9.59 & 10.68 & 8.95\\ \hline\hline
\end{tabular}
\end{center}
\label{table::coverage2}
\end{table}

\section{Exploring the Forest}
\label{section::explore}

In this section
we show how the forest guided smoother
can be used to examine properties of the forest.

A random forest is a complex object
and is difficult to interpret.
In contrast, the FGS is completely
determined by
the set of bandwidth matrices
$\Xi  =\{H_x: x\in \mathbb{R}^d\}$
which is a subset of the manifold of
symmetric positive-definite matrices.
We now consider a variety of methods
for summarizing and
exploring the set $\Xi$.
In this section we describe the methods.
Examples are given in Section \ref{section::examples}.

\subsection{Summarizing the Spatial Adaptivity of the Kernels}

Here we show how to
quantify the degree to which $H_x$
varies with $x$.
We take
$K$ to be a multivariate Gaussian.
The kernel at $x$ is
$K(x,H_x)$.
First we define
what the kernel looks like on average over $x$.
To do this we find the Wasserstein barycenter
of the distributions
$\{ K(0,H_{x})\}$.

The Wasserstein barycenter
comes from the theory of optimal transport;
a good reference on this area is
Peyr\'{e} and Cuturi (2019).
Recall first that the (second order) Wasserstein
distance between two distributions
$P_1$ and $P_2$
$$
W_2^2(P_1,P_2) =
\inf_J \mathbb{E}_J[ ||X-Y||^2]
$$
where
$X\sim P_1$,
$Y\sim P_2$
and the infimum is over all joint distributions $J$
with marginals $P_1$ and $P_2$.
In the special case of Normals, where
$P_1 = N(\mu_1,\Sigma_1)$ and
$P_2 = N(\mu_2,\Sigma_2)$ we have
$$
W^2(P_1,P_2) =
||\mu_1-\mu_2||^2 + 
{\rm tr}(\Sigma_1) + 
{\rm tr}(\Sigma_2) -2
{\rm tr}\Biggl\{  (\Sigma_1^{1/2} \Sigma_2 \Sigma_1^{1/2})^{1/2} \Biggr\}.
$$
The Wasserstein barycenter
of a set of distributions $Q_x$ indexed by $x$
is the distribution $\overline{Q}$
that minimizes
$$
\int W^2(Q_x,\overline{Q}) dP_X(x).
$$
This barycenter is useful because
it preserves the shape of the distributions.
For example,
the barycenter of
a $N(\mu_1,1)$ and 
$N(\mu_2,1)$ is
$N ( (\mu_1+\mu_2)/2,1)$.
The Euclidean average is the mixture
$(1/2) N(\mu_1,1) + (1/2) N(\mu_2,1)$
which does not preserve the shape of the
original densities.

In our case,
we summarize the set of bandwidth matrices
by finding the barycenter of the set of distributions
$\{K(0,H_{X_i})\}$.
The barycenter in this case can be
shown to be
$K(0,\overline{H})$
were
$\overline{H}$ is the unique positive definite matrix such that
\begin{equation}
\overline{H} =
\int (\overline{H}^{1/2} H_x \overline{H}^{1/2} )^{1/2} dP_X(x).
\end{equation}
In our examples,
we will compute
$\overline{H}$
to see what a typical bandwidth matrix looks like.
We also compute the Frechet variance
$$
V = \int W^2(\overline{H},H_x) dP_X(x)
$$
which gives a sense of how
much the bandwidth matrices vary over $x$.
If $H_{x}$ does not vary with $x$ then
then $V=0$.

Next we consider another way to summarize
the FGS.
For each $X_i$,
we find the effective bandwidth with respect to
each covariate
by finding the length of the ellipse 
$\{x:\ (u-X_i)^T H_{X_i}^{-1}(u-X_i)\leq c^2\}$
in the direction of each coordinate axis,
for any $c>0$.
In other words,
we compute
$\Delta_j(X_i) = \sqrt{c^2/H_{X_i}^{-1}(j,j)}$.
In the example section we'll see that
plots of
these quantities can be very informative.

\subsection{Comparing the Forest and the Smoother}
\label{section::compare}

How much prediction accuracy is
lost by using the smoother instead of the forest?
To answer this question we
define
$$
\Gamma = \E[ (Y-\hat\mu(X))^2 - (Y-\hat\mu_{RF}(X))^2].
$$
We can get an estimate of
$\Gamma$ using
the approach in 
\cite{williamson2020unified}.

Split the data into four groups
${\cal D}_1,{\cal D}_2,{\cal D}_3,{\cal D}_4$
each of size $m\approx n/4$.
From ${\cal D}_1$ get $\hat\mu_{RF}$ and from
${\cal D}_2$ get $\hat\mu$.
Let
$$
\hat\Gamma =
\frac{1}{m}\sum_{i\in {\cal D}_3} r_i -
\frac{1}{m}\sum_{i\in {\cal D}_4} s_i
$$
where
$$
r_i =  (Y_i - \hat\mu_{RF}(X_i))^2 ,\ \ \ 
s_i =  (Y_i - \hat\mu(X_i))^2.
$$
Then,
\cite{williamson2020unified}
show that
$$
\sqrt{m}(\hat{\Gamma}-\Gamma)\rightsquigarrow N(0,\tau^2)
$$
and a consistent estimate of $\tau^2$ is
$m^{-1}(\sum_i (r_i - \overline{r})^2 + \sum_i (s_i - \overline{s})^2)$.
Hence, a $1-\alpha$ confidence interval for $\Gamma$ is
$\hat{\Gamma} \pm z_{\alpha/2} \hat\tau/\sqrt{m}$.
(One can repeat this by permuting the blocks and averaging if desired.)

\subsection{Multiresolution Local Variable Importance}

One popular method
of assessing
local variable importance
is to estimate
the gradient of $\hat\mu$
or, equivalently, to use local
linear approximations
\cite{ribeiro2016should, plumb2018model}.
Using the forest guided local linear smoother
we get an estimate of the gradient and its
standard error for free. Furthermore, we can do this 
at various resolutions by varying $h$.

Let $\hat\beta_h(x) = (\hat\beta_{h,1}(x),\ldots, \hat\beta_{h,d}(x))$.
Now
$$
\hat\beta_{h,j}(x) = \sum_i Y_i \ell_{ij}(x; h H_x)
$$
where
$\ell_{ij}(x; h H_x)$ is the $i^{\rm th}$ element of the vector
$$
e_{j+1}^T (X_x^T W_x X_x)^{-1} X_x W_x,
$$
where
$W_{x}$ is a diagonal matrix with $W_{x}(i,i) = K(X_i- x;hH_x)$
and
$e_{j+1}$ is the vector that is all 0 except it is 1 in the
$j+1$ position. The standard error of $\hat\beta_{j,h}(x)$ is
${\rm se}_{j,h}(x) = 
\sqrt{ \sum_i \hat\sigma^2(X_i) \ell_{ij}^2(x;h H_x)}$.
A $1-\alpha$ variability interval is
$\hat\beta_{j,h}(x) \pm z_{\alpha/2} {\rm se}_{j,h}(x)$.

A plot of the values
$\hat\beta_{h,j}(X_i)$
gives a global sense of the local importance of the $j^{\rm th}$
covariate.
A plot of $\hat\beta_{h,j}(x)$
as a function of $h$ for a fixed $x$
summarizes local variable importance
at various resolutions.

\section{Examples}
\label{section::examples}

In this section,
we illustrate
the methods
from the previous section
on two examples.
The first is a synthetic example
and the second is a data example.

\vspace{-.5cm}
\subsection{Synthetic Example}

\begin{figure}
\begin{center}
\includegraphics[scale=.45]{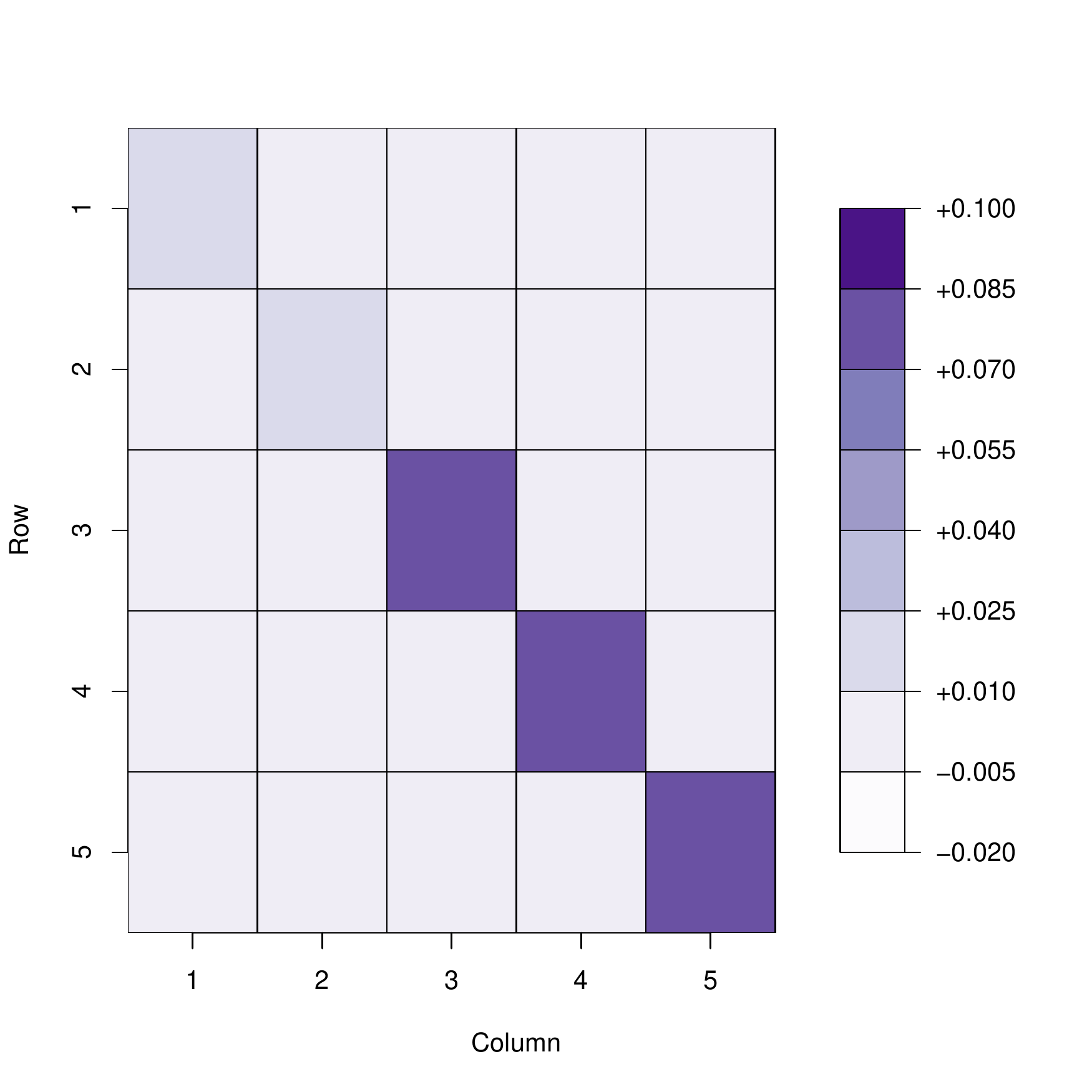}
\end{center}
\vspace{-.5cm}
\caption{Barycenter of the bandwidth matrices
for the example in equation (\ref{eq::mu2}).}
\label{fig::Stefan_Bary}
\end{figure}

We return to the example given in (\ref{eq::mu2}).
Figure \ref{fig::Stefan_Bary}
shows the Wasserstein barycenter of the bandwidth matrices.
The barycenter shows that the typical bandwidth for the 
first two variables is small. This makes sense as the function 
only depends on $x_1$ and $x_2$. Also, the small off-diagonals 
suggest the bandwidth matrix is typically not far from diagonal.

\newpage

The Frechet variance is 0.019, suggesting that the bandwidth 
matrix does not vary greatly across the sample space.

\begin{figure}
\begin{center}
\includegraphics[width=4in,height=3in] 
        {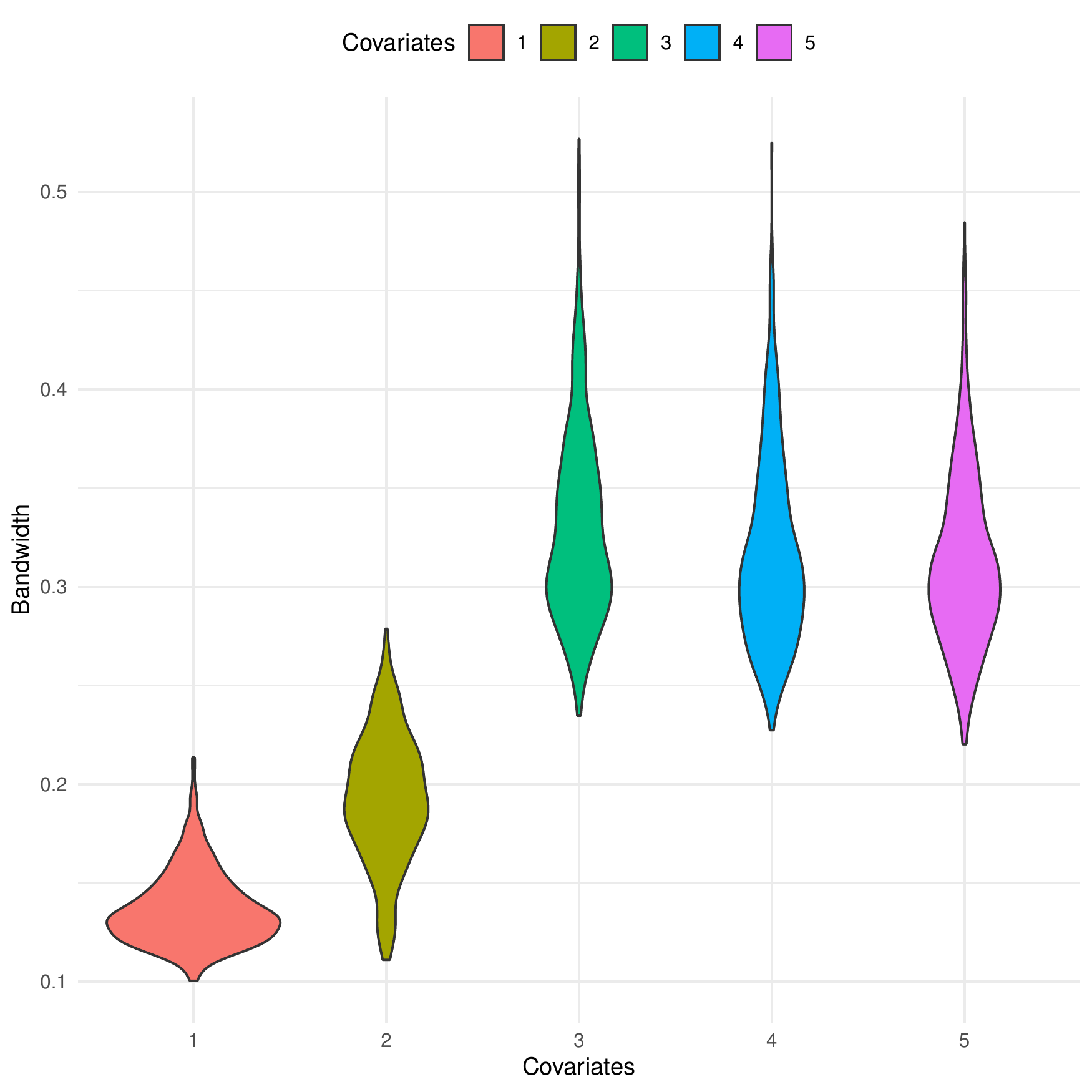}
\end{center}
\vspace{-.5cm}
\caption{Example \eqref{eq::mu2}. Effective bandwidths 
for each covariate.}
\label{fig::violins1}
\end{figure}

Figure \ref{fig::violins1} shows violin plots of effective 
bandwidths. The effective bandwidth $\Delta_j$
is smaller for $x_1$ and $x_2$ than for the other variables.
This is what we would expect since $\mu(x)$ does not depend
on $x_3,x_4$ or $x_5$.
The forest attempts to smooth over these irrelevant variables
and hence the approximating bandwidth matrices tend to be large 
in the directions
of the irrelevant variables.
This confirms what we found with the barycenter.

\begin{figure}
\begin{center}
\includegraphics[width=5.5in,height=2.8in]
        {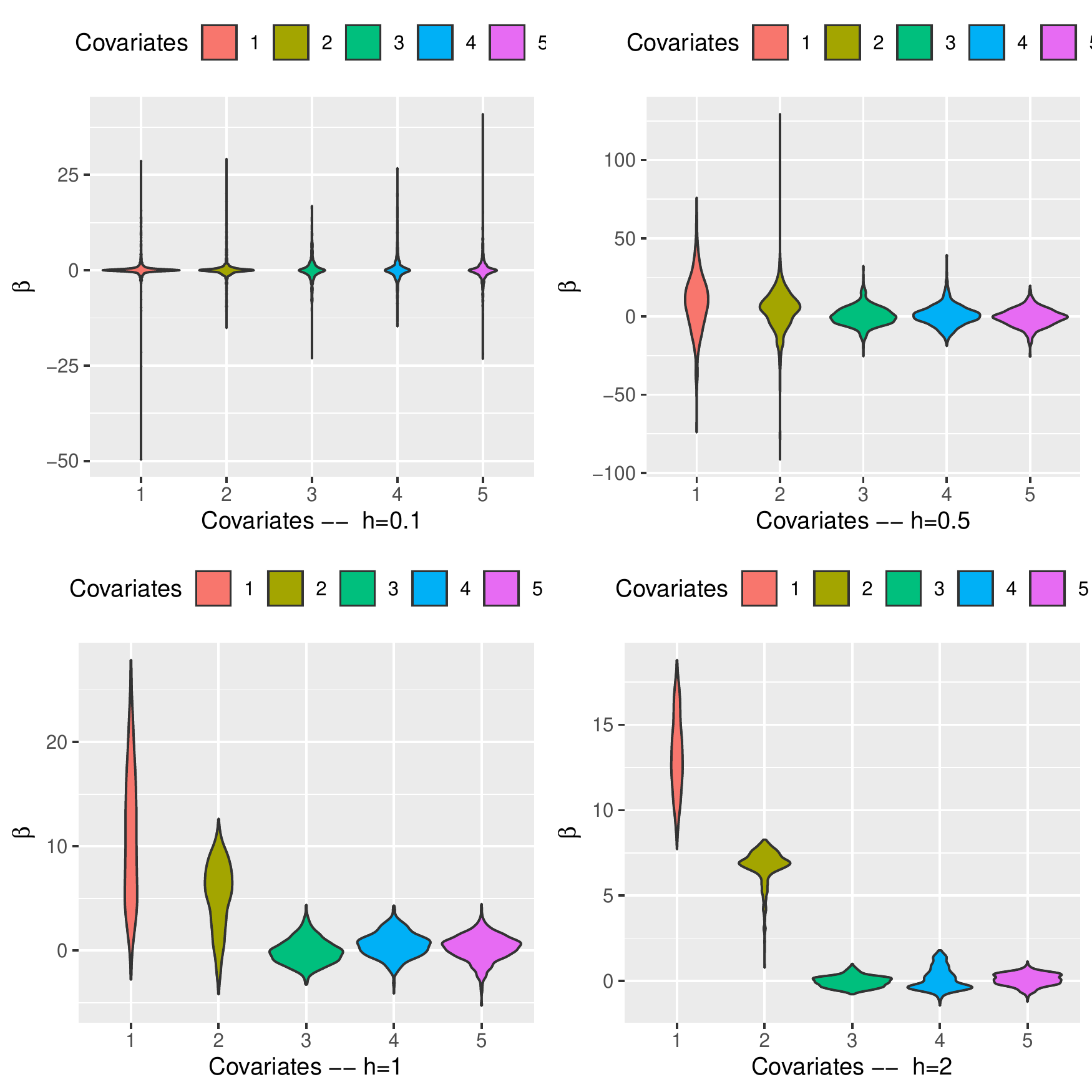}
\end{center}
\vspace{-.5cm}
\caption{$\{\hat\beta_j(X_1),\ldots, \hat\beta_j(X_n)\}$ for 
each covariate at four resolutions,
$h=0.1, 0.5, 1, 2$.}
\label{fig::violins2}
\end{figure}

The four plots in Figure \ref{fig::violins2}
show the local slopes 
$\hat\beta_1(X_i),\ldots,\hat\beta_5(X_i)$
for each of the five covariates (over all $X_i$)
at four different resolutions. 
and variable importance (bottom)
at several resolutions $h=0.1, 0.5, 1$ and $2$.
The two smallest resolutions ($h=0.1, h=0.5$) are uninformative.
The two larger resolutions ($h=1$, $h=2$)
provide clear evidence of the importance of $x_1$ and $x_2$.
Note that importance variables correspond to small bandwidths 
but large slopes.

\begin{figure}
\begin{center}
\includegraphics[width=5.5in,height=2.8in]
        {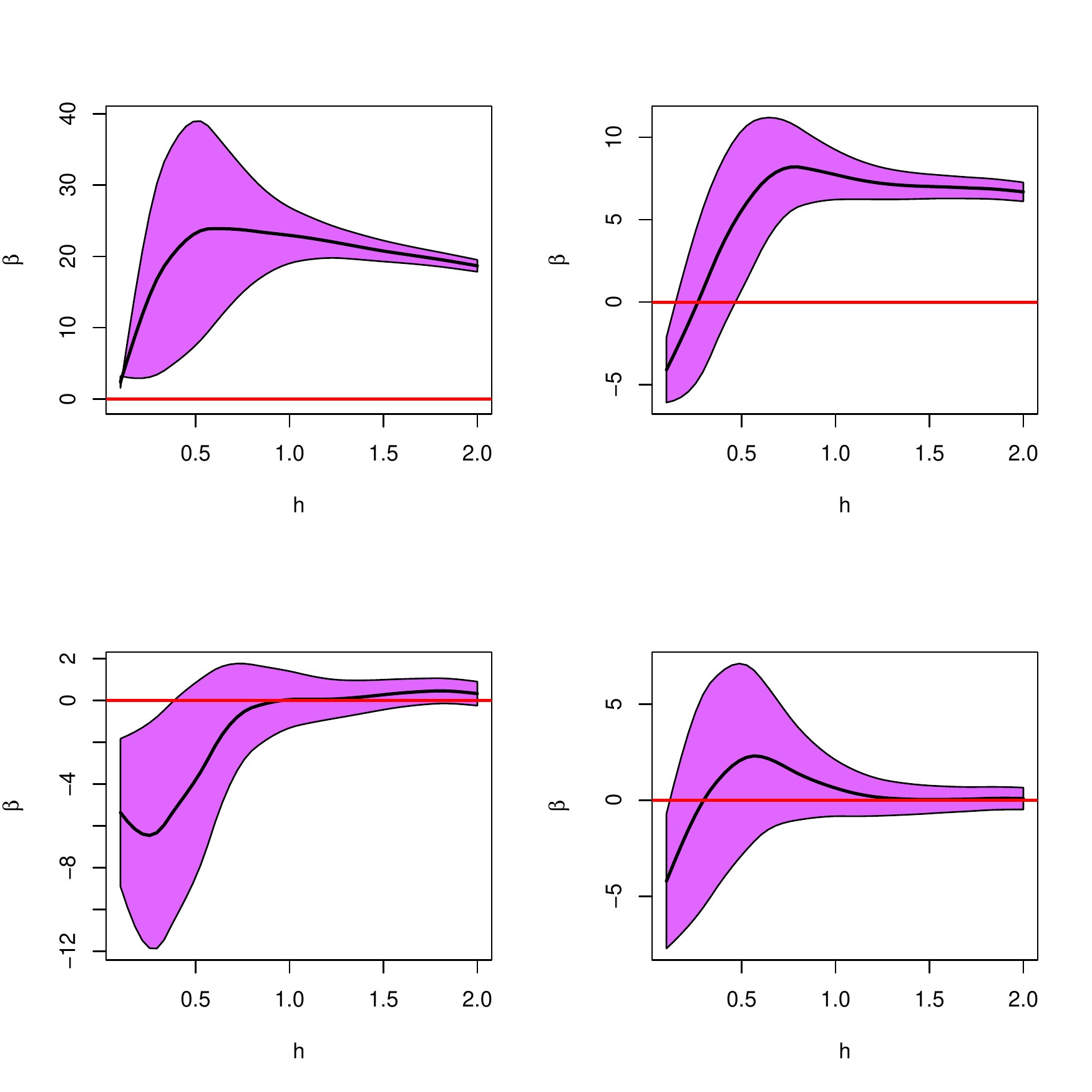}
\end{center}
\vspace{-.5cm}
\caption{Variability intervals for 
$\hat\beta_{1,h}(x), \cdots,\hat\beta_{4,h}(x)$ at
$x=(1/2,1/2,1/2,1/2,1/2)$. }
\label{fig::varimp}
\end{figure}

Figure \ref{fig::varimp} shows variability intervals for 
$\hat\beta_{1,h}(x), \cdots,\hat\beta_{4,h}(x)$ at\\
$x=(1/2,1/2,1/2,1/2,1/2)$, the center of the support of 
$X$. (The fifth variable is not shown.)
These intervals are plotted versus increasing values of $h$
resulting in (pointwise) variability bands for
$\beta_{j,h}(x)$.

Again, we see that $x_1$ and $x_2$
are the important variables
as the bands exclude 0 for larger values of $h$
while the bands for $x_3$ and $x_4$
include 0 for all $h$.

\begin{figure}
\begin{center}
\includegraphics[scale=.56]{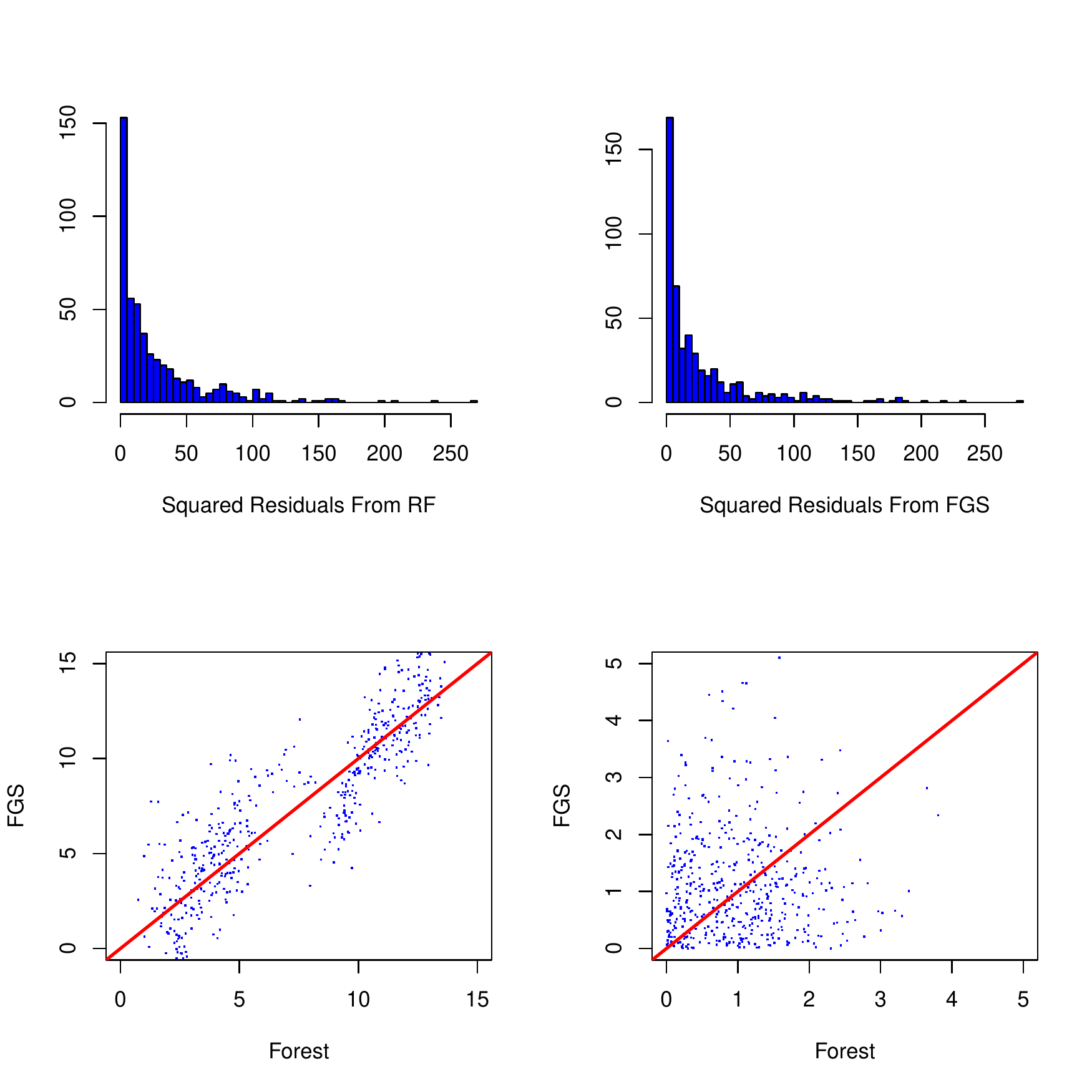}
\end{center}
\vspace{-.6cm}
\caption{\small Top left: squared residuals from the 
random forest.
Top right: squared residuals from the FGS.
Bottom left: Plot of $\hat\mu_{RF}(X_i)$ 
versus $\hat\mu(X_i)$.
Bottom right: Plot of
$|Y_i - \hat{\mu}_{RF}(X_i)|$ versus 
$|Y_i - \hat{\mu}(X_i)|$.}
\label{fig::plot_forest_vs_fgs_stefan}
\end{figure}

Next we compare the FGS to the forest.
The top left plot
of Figure \ref{fig::plot_forest_vs_fgs_stefan}
shows histograms of the squared residuals
for the forest and of the FGS.
The two histograms are very similar.
It also shows two scatterplots of
$\hat{\mu}_{RF}(X_i)$ and e$\hat{\mu}(X_i)$.
and of their residuals.

\bigskip
We do see a very slight loss in accuracy for the FGS but 
the difference is small.
It appears that  the two fits
are very similar.
To formalize this,
we estimate $\Gamma$ as described in Section 
\ref{section::compare}
and we find that the 95 per cent confidence interval
$\hat\Gamma = -0.134 \pm 4.9$
again suggesting little difference between the two methods.
Thus we conclude that the FGS appears to be a good 
approximation to the forest.

\subsection{Covid-19}

In this section
we consider data
on Covid-19
obtained from the API 
of the CMU Delphi group at
\url{covidcast.cmu.edu}.

Our goal is to construct a random forest to predict
$Y= $ average daily deaths
from these variables:

\begin{enumerate} 
\item[]
cli \hspace{3em}
Percentage of people with Covid-like symptoms 
(surveys of~Facebook~ users)
\vspace{-1.3cm}

\item [] 
dr \hspace{3em} Percentage of daily doctor visits that are 
due to Covid-like symptoms
\vspace{-.5cm}

\item[]
cases \hspace{1.7em} Newly reported Covid-19 cases per 100,000 
people
\vspace{-.5cm}

\item [] 
home \hspace{1.6em} Proportion of people staying home
\vspace{-.5cm}  

\item [] 
masks \hspace{1.2em}  Percentage of people who say they 
wear a mask in public
\vspace{-.5cm}  

\item [] 
hospital \hspace{.4em}  Percentage of daily hospital 
admissions with Covid-19
\vspace{-.5cm}  

\item [] 
prevdeaths \hspace{.3em} Previous number of deaths due to Covid-19

\end{enumerate}

The variable $Y$ is averaged over December 1 2020 to 
December 12 2020.
The covariates are averaged from
October 1 2020 to December 1 2020.
We took the logarithms of all variables
and then scaled each covariate to
have mean 0 and variance 1.

The problem of predicting the epidemic
is an intensely studied issue
and our goal is not to
develop a cutting edge prediction method.
Rather, we use these data as a vehicle for
illustrating our methods.

After fitting the FGS
we can summarize the local fit
for various counties
by reporting
the local slopes
$\hat\beta_1(x),\ldots,\hat\beta_d(x)$
and their standard errors.
Table 1 and Table 2, below 
show this for four counties.
The nice thing about the FGS is that
we can describe the model for any county
in the familiar form of a (local) linear model.
This makes the model very interpretable
for users such as public health officials.

\medskip
\begin{tabular}{cc}
\begin{tabular}{|lcc|}
\multicolumn{3}{c}{\bf New York County, NY}\\ 
\hline \hline\\
\vspace{-1.2cm}\\
Coefficients & $\hat\beta$ & Standard Error\\ \hline
\vspace{-.3cm}
cli &  -0.026 &  0.057\\
\vspace{-.3cm}
dr  &  -0.042 &  0.034\\
\vspace{-.3cm}
cases & -0.012 & 0.044\\
\vspace{-.3cm}
home  & 0.148 & 0.056\\
\vspace{-.3cm}
masks & -0.016 &  0.042\\
\vspace{-.3cm}
hospital & 0.003 & 0.058\\
prevdeaths & 0.128 & 0.082\\ \hline
\end{tabular} 
\end{tabular} 
\begin{tabular}{cc}
\begin{tabular}{|lcc|}
\multicolumn{3}{c}{\bf Elkhart County, IN}\\ 
\hline \hline\\
\vspace{-1.2cm}\\
Coefficients & $\hat\beta$ & Standard Error\\ \hline
\vspace{-.3cm}
cli &  0.015 &  0.119\\
\vspace{-.3cm}
dr  &  0.155 &  0.148\\
\vspace{-.3cm}
cases & -0.150 & 0.106\\
\vspace{-.3cm}
home  & 0.204 & 0.159\\
\vspace{-.3cm}
masks & 0.124 &  0.118\\
\vspace{-.3cm}
hospital & -0.158 & 0.098\\
prevdeaths & 0.141 & 0.087\\ \hline
\end{tabular} 
\end{tabular} 

\medskip
\begin{center}
{\bf Table 3}
\end{center}

\medskip

\begin{tabular}{cc}
\begin{tabular}{|lcc|}
\multicolumn{3}{c}{\bf DuPage County, IL}\\ 
\hline \hline\\
\vspace{-1.2cm}\\
Coefficients & $\hat\beta$ & Standard Error\\ \hline
\vspace{-.3cm}
cli &  0.62 &  0.077\\
\vspace{-.3cm}
dr  &  0.032 &  0.049\\
\vspace{-.3cm}
cases & -0.029 & 0.063\\
\vspace{-.3cm}
home  & -0.006 & 0.089\\
\vspace{-.3cm}
masks & 0.013 &  0.083\\
\vspace{-.3cm}
hospital & -0.061 & 0.100\\
prevdeaths & 0.211 & 0.069\\ \hline
\end{tabular} 
\end{tabular} 
\hspace{.2cm}
\begin{tabular}{|lcc|}
\multicolumn{3}{c}{\underline{Lubbock County, TX}}\\ 
\hline \hline\\
\vspace{-1.2cm}\\
Coefficients & $\hat\beta$ & Standard Error\\ \hline
\vspace{-.3cm}
cli &  0.043 &  0.094\\
\vspace{-.3cm}
dr  &  0.110 &  0.097\\
\vspace{-.3cm}
cases & -0.066 & 0.080\\
\vspace{-.3cm}
home  & 0.105 & 0.108\\
\vspace{-.3cm}
masks & 0.063 &  0.072\\
\vspace{-.3cm}
hospital & -0.065 & 0.061\\
prevdeaths & 0.107 & 0.074\\ \hline
\end{tabular} 

\smallskip
\begin{center} 
{\bf Table 4}
\end{center}

\begin{figure}
\begin{center}
\begin{tabular}{c}
\includegraphics[width=5in,height=3in]{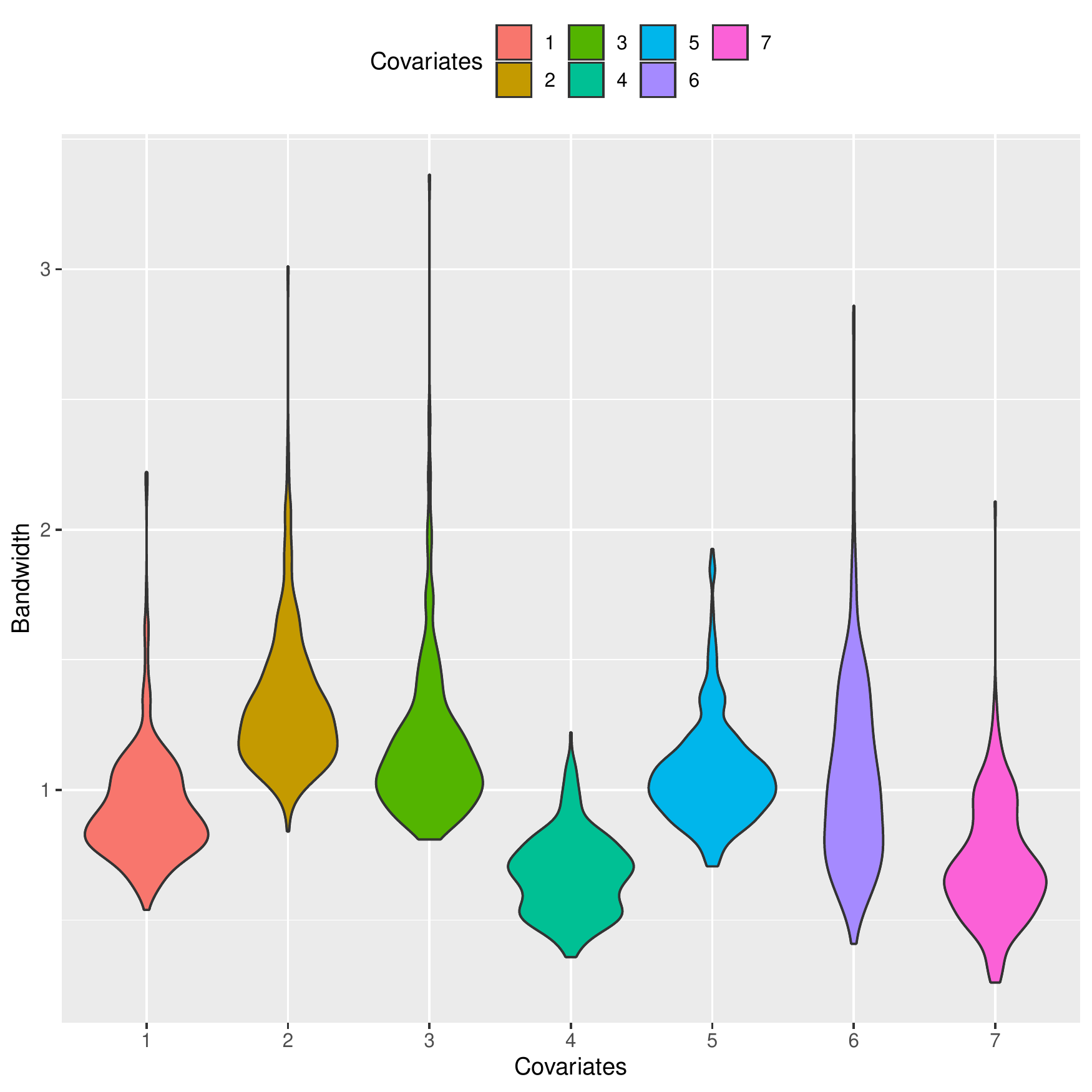} \\
\includegraphics[width=5in,height=3in]{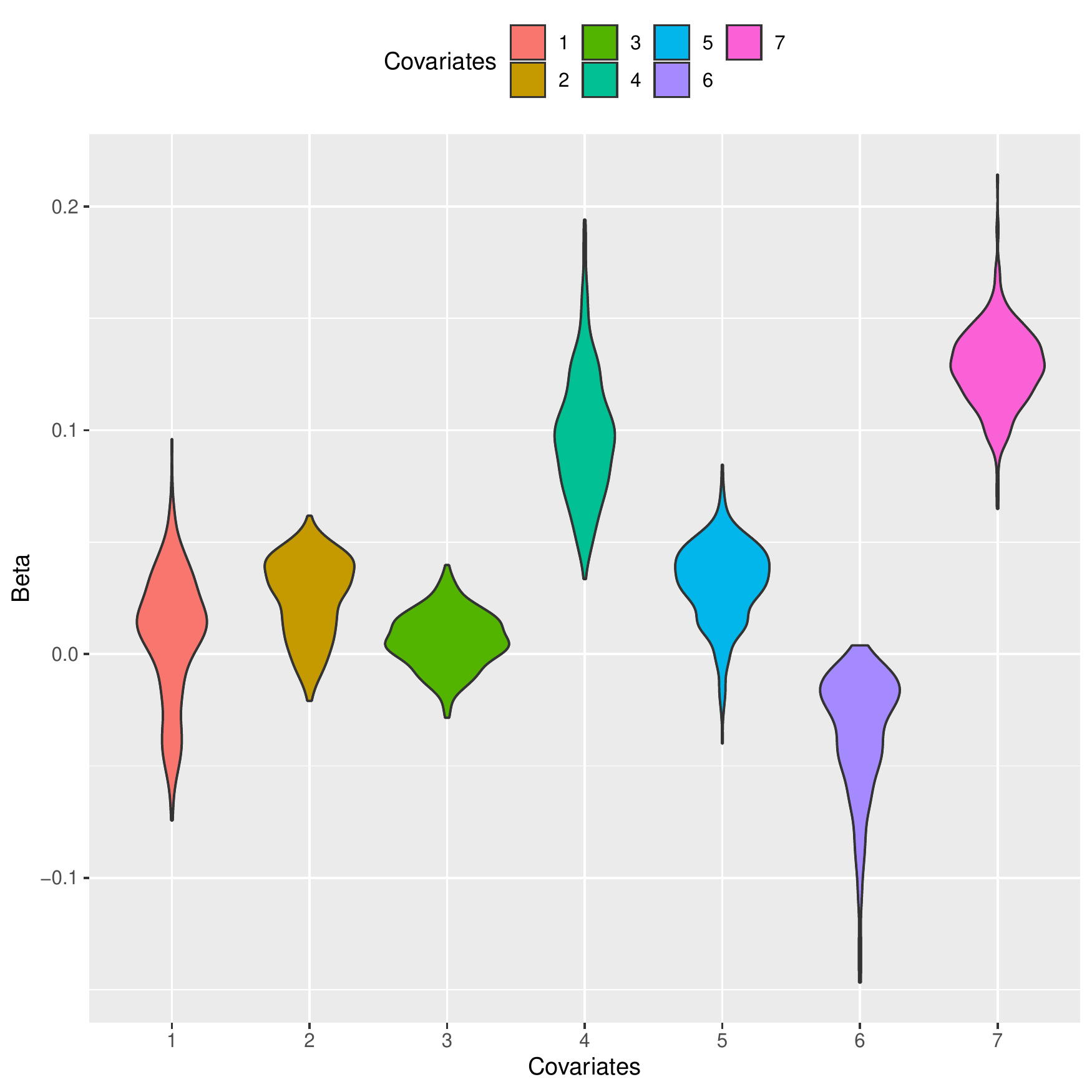}
\end{tabular}
\end{center}
\caption{Bandwidth plot (top) and $\beta$ plot (bottom)
for Covid example}
\label{fig::covid-bandwidth}
\end{figure}

Figure \ref{fig::covid-bandwidth}
shows the effective bandwidths and
local slopes at resolution $h=2$.
The two most important variables
(small bandwidths and large slopes)
are $x_4$ (home) and $x_7$ (previous deaths).
The importance of previous deaths is obvious.
The fact that social mobility (home) is important
is notable but we should emphasize that this is a
predictive analysis not a causal analysis.

\begin{figure}
\begin{center}
\includegraphics[scale=.5]{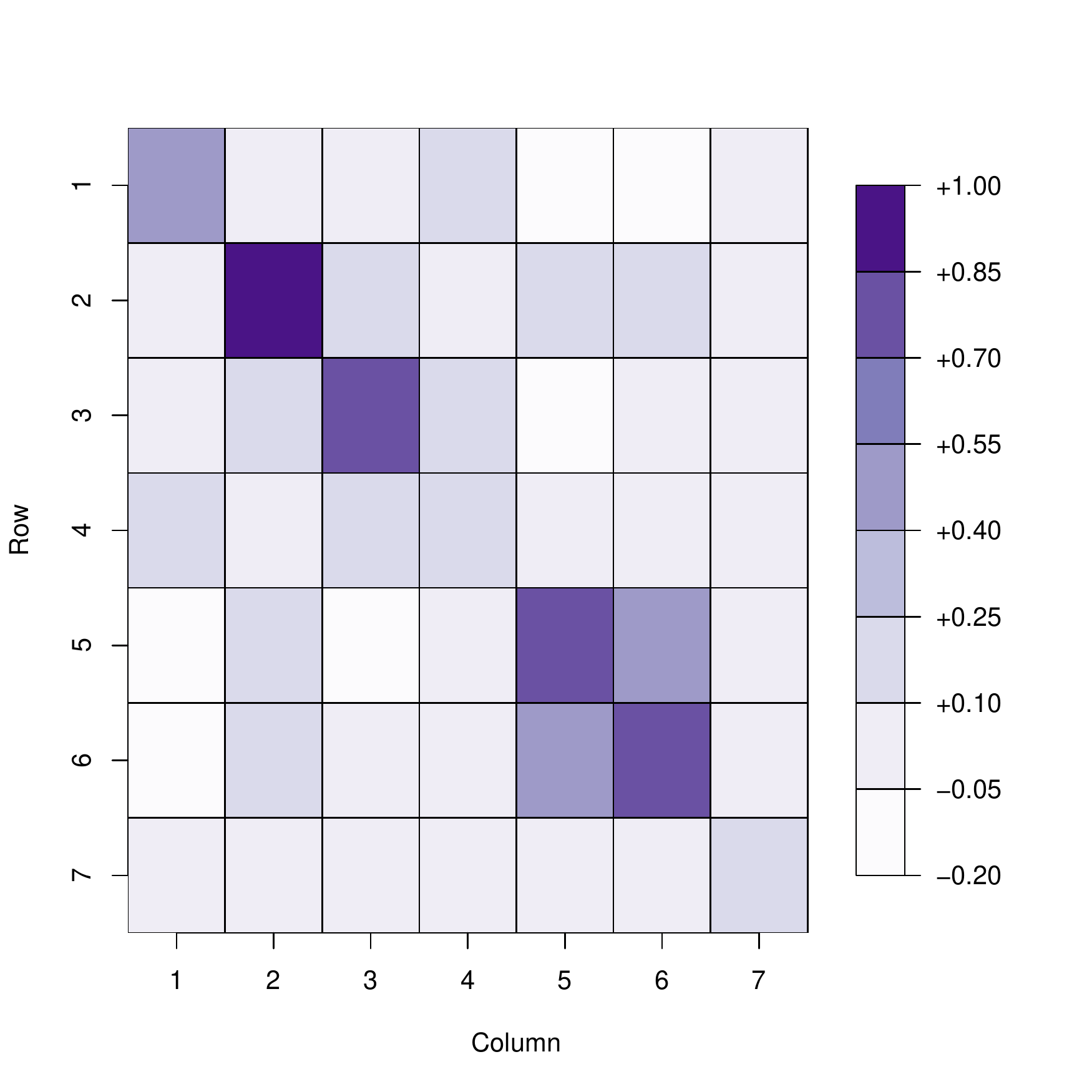}
\end{center}
\caption{Barycenter of the bandwidth matrices for 
Covid example.}
\label{fig::covid-bary}
\end{figure}

Figure \ref{fig::covid-bary} shows the barycenter of the 
bandwidth matrices.
Note that the fourth and seventh elements on the diagonal 
are the smallest confirming the importance of those variables.
We also see some correlation between the bandwidths for $x_5$ 
and $x_6$. The Frechet variance is 0.817
suggesting that $H_x$ varies quite a bit with $x$
(recall that all the variables are scaled to have variance 1).

\section{Discussion}
\label{section::discussion}

Throughout this paper we have
assumed that the number of covariates $d$ is fixed.
If $d$ increases with $n$
then local linear fitting will not
work.
Instead one will need to include
some sort of ridge or $\ell_1$ penalty.
Furthermore, when $d$ is large,
$H_x$ will not be invertible
and so regularization on $H_x$
is required.

We have focused on random forests
but similar ideas can be used for other
black box methods
such as neural nets.
\cite{koh2017understanding}
show how to compute the
influence function
for deep nets and other predictors.
The influence function
can be used to define
a spatially adaptive kernel
as we have done using the weights
from a forest.

In our examples
we have not found
much difference
between the forest and the FGS.
But this may be due to the fact
that we have not considered
complex high dimensional problems.
Understanding when a complex predictor
can be approximated by a spatially varying 
local smoother
is a interesting but challenging problem.

\bibliographystyle{paper}
\bibliography{Bibliography-MM-MC}

\section*{Appendix}

Here we recall Theorem
\ref{thm::clt}
and give an outline of the proof.

{\bf Theorem 3.1}
Assume that
(i) $\sup_x | \hat\sigma^2(x) - \sigma^2(x)| \stackrel{P}{\to} 0$,
(ii) $\sigma^2(x)>0$,
$$
(iii)\ \ \ \ -a < \gamma < \frac{1-ad}{d}
$$
and further, if
$t < d/2$ we require
$a < 1/(d-2t)$.
Also, assume that $Y$ is bounded and that $b > t+1$.
Then
$$
\frac{\mu^\dagger(x) - \mu(x)}{\sqrt{\hat{\rm Var}[\mu^\dagger(x)]}}
                     \rightsquigarrow N(0,1).
$$

{\sc Proof Outline.}
First note that the condition
$\gamma > -a$ ensures that
$n|h H_x|\to \infty$
and this implies
${\rm Var}[\mu^\dagger(x)]\to 0$.
We write
$$
\frac{\mu^\dagger(x) - \mu(x)}{\sqrt{{\rm Var}(\mu^\dagger(x))}} =
\frac{\mu^\dagger(x) - \E[\mu^\dagger(x)]}{\sqrt{{\rm Var}(\mu^\dagger(x))}} +
\frac{\E[\mu^\dagger(x)] - \mu(x)}{\sqrt{{\rm Var}(\mu^\dagger(x))}}.
$$
Recall that
$\mu^\dagger(x) = e_1^T ( {\cal H}^T {\cal H} )^{-1} {\cal H}^T \hat m$
where
$\hat m = (\hat \mu(x; h_1 H_x), \ldots, \hat \mu(x; h_b H_x))^T$.
Now
$\E[\hat m] = {\cal H} \kappa_n(x) + o(n^{-at})$
where we recall that
$$
\kappa_n(x) = \Bigl(\mu(x), \sum_{j=2}^t c_j(x) h_1^j/n^{aj},\ldots, 
              \sum_{j=2}^t c_j(x) h_b^j/n^{ja}\Bigr)^T.
$$
Hence,
$$
\E[\mu^\dagger(x) - \mu(x)] =
e_1^T ( {\cal H}^T {\cal H} )^{-1} {\cal H}^T [{\cal H} \kappa_n(x) + 
                  o(n^{-at})] =o(n^{-at}).
$$
Let
${\cal V}$ be the covariance matrix of $\hat m$.
Then, arguing as in the proof of Theorem 2.1 of
\cite{ruppert1994multivariate},
there exists a $b\times b$ positive definite matrix
$A$ depending on $K$,
$\alpha_1,\ldots,\alpha_b,x,f(x)$ and $\sigma^2(x)$
but not on $n$, such that
$$
{\cal V} = \frac{A}{n |h_1 H_x|} (1+o_P(1)).
$$
Hence,
\begin{align*}
{\rm Var}[\mu^\dagger(x)] &=
e_1^T ( {\cal H}^T {\cal H} )^{-1} {\cal H}^T {\cal V} {\cal H} 
                     ( {\cal H}^T {\cal H} )^{-1} e_1\\
&=
\frac{1}{  n | h_1 H_x|} e_1^T ({\cal H}^T {\cal H})^{-1} e_1 (1+o_P(1))\\
&=
\frac{1}{  n | h_1 H_x|}   (1+o_P(1))\\
&= O_P( n^{1-d(a+\gamma)})
\end{align*}
since
$e_1^T ({\cal H}^T {\cal H})^{-1} e_1 =O(1)$.
Since
$\gamma < (1-ad)/d \leq \min\{ (1-ad)/d, (2at-da+1)/d\}$
it follows that
$$
\frac{\E[\mu^\dagger(x)] - \mu(x)}{\sqrt{{\rm Var}(\mu^\dagger(x))}} = o_P(1).
$$
Now
$\E[ |\mu^\dagger(x) - \E[\mu^\dagger(x)]|^3= O(n^{2d(a+\gamma)-2})$.
Hence
$$
\frac{\E[ |\mu^\dagger(x) - \E[\mu^\dagger(x)]|^3]}
{{\rm Var}[\mu^\dagger(x)]^{3/2}} =
O (n^{2d(a+\gamma)-2} n^{(3/2)(1-d(a+\gamma))}) = o_P(1)
$$
since
$\gamma < (1-ad)/d$.
Hence, by Lyapunov's central limit theorem,
$$
\frac{\mu^\dagger(x) - \E[\mu^\dagger(x)]}{\sqrt{{\rm Var}(\mu^\dagger(x))}} 
                     \rightsquigarrow N(0,1).
$$
Finally,
since $\sup_x | \hat\sigma^2(x) - \sigma^2(x)| \stackrel{P}{\to} 0$ and
$\sigma^2(x)>0$,
it follows that
$\hat{\rm Var}[\mu^\dagger(x)]/{\rm Var}[\mu^\dagger(x)] \stackrel{P}{\to} 1$
and the result follows.

\end{document}